\documentclass[10pt,twocolumn,letterpaper]{article}

\usepackage{cvpr}
\usepackage{times}
\usepackage{epsfig}
\usepackage{graphicx}
\usepackage{amsmath}
\usepackage{amssymb}

\usepackage{makecell}
\usepackage{caption}
\usepackage[pagebackref=true,breaklinks=true,letterpaper=true,colorlinks,bookmarks=false]{hyperref}

\cvprfinalcopy % *** Uncomment this line for the final submission

\def\cvprPaperID{6410} % *** Enter the CVPR Paper ID here

\ifcvprfinal\fi
\begin{document}

%%%%%%%%% TITLE
\title{Spatio-Temporal Graph for Video Captioning with Knowledge Distillation}
\author{Boxiao Pan$^1$, Haoye Cai$^1$, De-An Huang$^1$, \\ Kuan-Hui Lee$^2$, Adrien Gaidon$^2$, Ehsan Adeli$^1$, Juan Carlos Niebles$^1$ \\
{$^1$Stanford University \quad $^2$Toyota Research Institute} \\
{\tt\small \{bxpan,hcaiaa,dahuang,eadeli,jniebles\}@cs.stanford.edu \tt\small \{kuan.lee,adrien.gaidon\}@tri.global}
}

\twocolumn[{%
\renewcommand\twocolumn[1][]{#1}%
\maketitle
\begin{center}
    \centering
    \includegraphics[width=\linewidth]{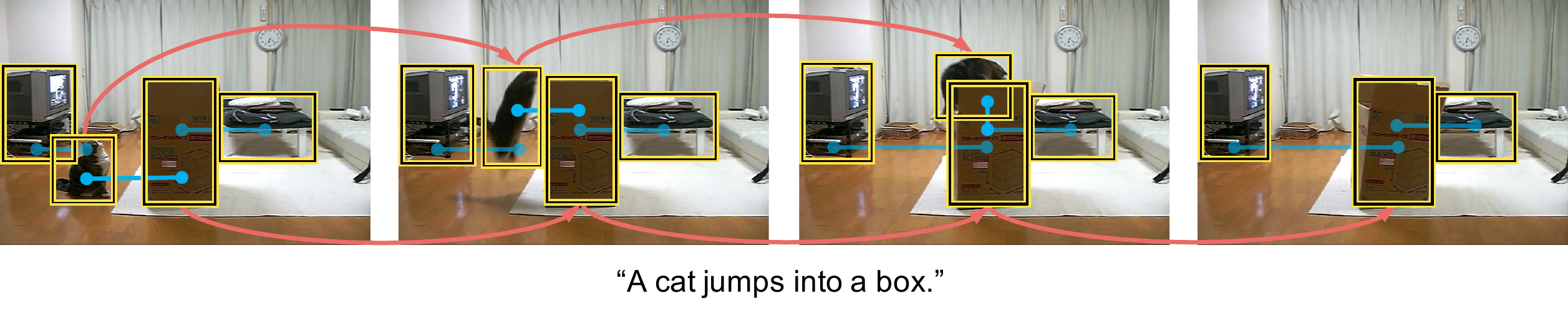}
    \captionof{figure}{How to understand and describe a scene from video input? We argue that a detailed understanding of spatio-temporal object interaction is crucial for this task. In this paper, we propose a spatio-temporal graph model to explicitly capture such information for video captioning. Yellow boxes represent object proposals from Faster R-CNN \cite{ICCV2015FasterRCNN}. Red arrows denote directed temporal edges (for clarity, only the most relevant ones are shown), while blue lines indicate undirected spatial connections. Video sample from MSVD \cite{ACL2011MSVD} with the caption ``A cat jumps into a box." Best viewed in color.}
    \label{fig:pull_fig}
\end{center}%
}]

%\maketitle
%\thispagestyle{empty}

%%%%%%%%% ABSTRACT
\begin{abstract}

Video captioning is a challenging task that requires a deep understanding of visual scenes. State-of-the-art methods generate captions using either scene-level or object-level information but without explicitly modeling object interactions. Thus, they often fail to make visually grounded predictions, and are sensitive to spurious correlations. In this paper, we propose a novel spatio-temporal graph model for video captioning that exploits object interactions in space and time. Our model builds interpretable links and is able to provide explicit visual grounding. To avoid unstable performance caused by the variable number of objects, we further propose an object-aware knowledge distillation mechanism, in which local object information is used to regularize global scene features. We demonstrate the efficacy of our approach through extensive experiments on two benchmarks, showing our approach yields competitive performance with interpretable predictions.

\end{abstract}

%%%%%%%%% BODY TEXT
\section{Introduction}

Scenes are complicated, not only because of the diverse set of entities involved, but also the complex interactions among them. Consider the scene shown in Fig.~\ref{fig:pull_fig}. In order to understand that ``A cat jumps into a box," we need to first identify ``cat" and ``box," then capture the transformation of ``cat jumps into the box." It is also crucial to be able to ignore the ``television" and ``bed," since they mostly serve as distractors for understanding what is happening. 

The task of video captioning \cite{ICCV2013Youtube2Text, ICCV2015S2VT} approaches scene understanding by generating text descriptions from video input. However, current methods for video captioning are not able to capture these interactions. Rather than modeling the correlations among high-level semantic entities, current methods build connections directly on raw pixels and rely on the hierarchical deep neural network structure to capture higher-level relationships \cite{ICCV2019JointSyntaxVideoCap, Arxiv2019ControllableVideoCap}. Some works try operating on object features instead, but they either ignore cross-object interaction \cite{CVPR2019OABTG}, or object transformation over time \cite{CVPR2018AttendNInteract, CVPR2019GroundedVideoCap}. Despite efforts in directly modeling local object features, the connections among them are not interpretable \cite{CVPR2018AttendNInteract, CVPR2019GroundedVideoCap}, and hence sensitive to spurious correlations.

On the other hand, modeling object relations via video spatio-temporal graphs \cite{CVPR2019GatedSTGraph, ECCV2018STRegionGraph} has been explored to explicitly construct links between high-level entities by leveraging the relation-modeling nature of graphs. Specifically, nodes represent these entities, such as body joints \cite{AAAI2018STGCN}, objects / persons \cite{Arxiv2019GazeSTGraph, ECCV2018STRegionGraph, CVPR2019GroupActivityGraph}, and actions \cite{CVPR2019GatedSTGraph}, while edges encode relationships among the entities. Although spatio-temporal graph models have achieved great success on classification tasks \cite{Arxiv2019GazeSTGraph, Arxiv2018CollisionSTGraph, ECCV2018STRegionGraph, CVPR2019GroupActivityGraph}, the effect of relation modeling remains unclear, as the model would easily shortcut the classification problem by taking advantage of other cues (e.g., background). To the best of our knowledge, we are the first to explicitly model spatio-temporal object relationships for video captioning, and show the effect of graphical modeling through extensive experiments.

To provide the global context that is missing from local object features, previous work either merges them to another global scene branch through feature concatenation \cite{ECCV2018STRegionGraph} or pooling \cite{CVPR2019OABTG}, or adds scene features as a separate node in the graph \cite{Arxiv2019GazeSTGraph, Arxiv2018StackedSTGraph, CVPR2019GatedSTGraph}. However, because videos contain a variable number of objects, the learned object representation is often noisy. It thus leads to suboptimal performance. To solve this problem, we introduce a two-branch network structure, where an object branch captures object interaction as privileged information, and then injects it into a scene branch by performing knowledge distillation \cite{Arxiv2015KnowledgeDistill} between their language logits. Compared with previous approaches that impose hard constraints on features, our proposed method applies soft regularization on logits, which thus makes the learned features more robust. We refer to this mechanism as ``object-aware knowledge distillation." During testing, only the scene branch is used, which leverages the distilled features with object information already embedded. As a bonus effect, this approach is also able to save the cost of running object detection at test time.

In this paper, we propose a novel way to tackle video captioning by exploiting the spatio-temporal interaction and transformation of objects. Specifically, we first represent the input video as a spatio-temporal graph, where nodes represent objects and edges measure correlations among them. In order to build interpretable and meaningful connections, we design the adjacency matrices to explicitly incorporate prior knowledge on the spatial layout as well as the temporal transformation. Subsequently, we perform graph convolution \cite{Arxiv2016GCN} to update the graph representation. This updated representation is then injected into another scene branch, where we directly model the global frame sequences, as privileged object information via the proposed object-aware knowledge distillation mechanism. Afterward, language decoding is performed through a Transformer network \cite{NIPS2017Transformer} to obtain the final text description. We conduct experiments on two challenging video captioning datasets, namely MSR-VTT \cite{CVPR2016MSRVTT} and MSVD \cite{ACL2011MSVD}. Our model demonstrates significant improvement over state-of-the-art approaches across multiple evaluation metrics on MSVD and competitive results on MSR-VTT. Note that although our proposed model is agnostic to downstream tasks, we only focus on video captioning in this work. Its application on other domains is thus left as future work.

In summary, our \textbf{main contributions} are as follows. (1) We design a \textbf{novel spatio-temporal graph network} to perform video captioning by exploiting object interactions. To the best of our knowledge, this is the first time that spatio-temporal object interaction is explicitly leveraged for video captioning and in an interpretable manner. (2) We propose an \textbf{object-aware knowledge distillation mechanism} to solve the problem of noisy feature learning that exists in previous spatio-temporal graph models. Experimental results show that our approach achieves a significant boost over the state-of-the-art on MSVD~\cite{ACL2011MSVD} and competitive results on MSR-VTT~\cite{CVPR2016MSRVTT}.

\begin{figure*}[t]
\begin{center}
\includegraphics[width=.9\linewidth]{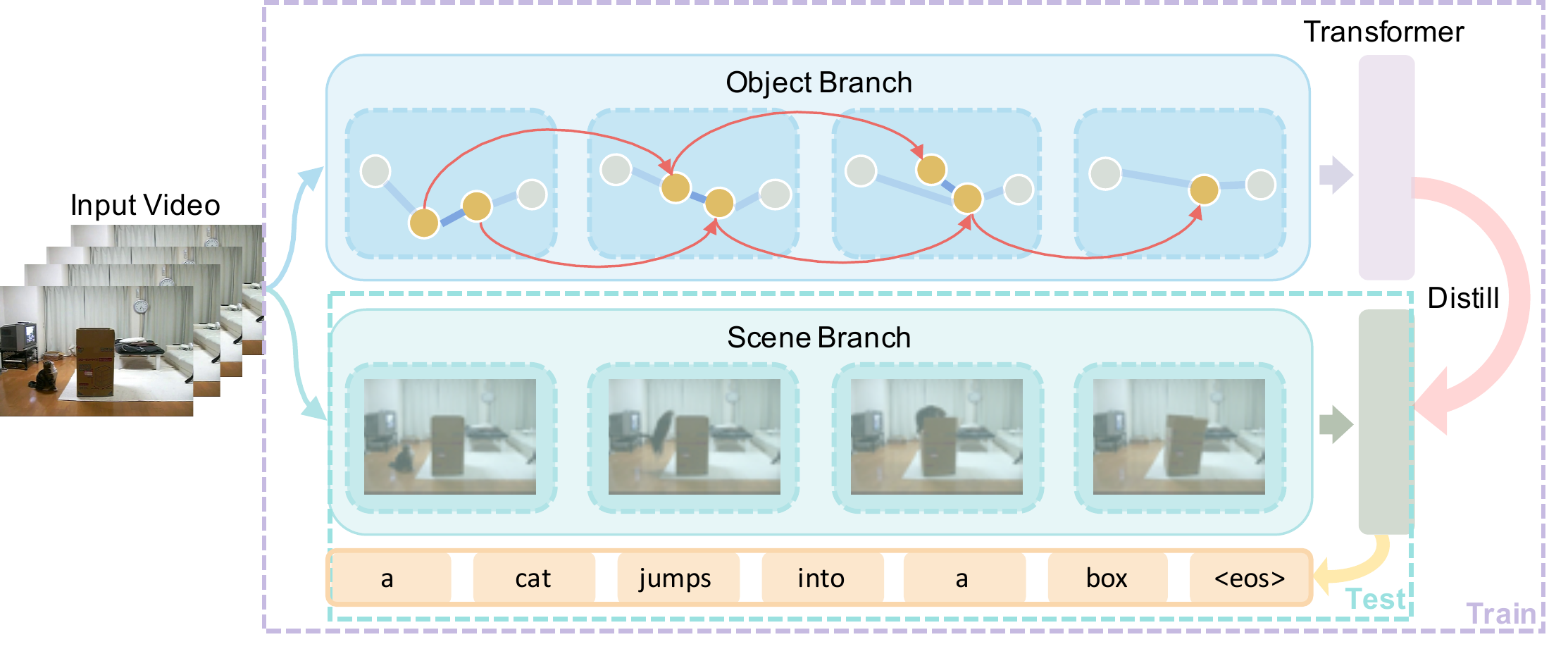}
\end{center}
   \caption{Overview of the proposed two-branch framework. During training, the object branch captures space-time object interaction information via the proposed spatio-temporal graph model, while the scene branch provides the global context absent from the object branch. The object-level information is then distilled into the scene feature representation by aligning language logits from the two branches. For clarity, we drop the arrow from the object branch Transformer to the output sentence, but it is also trained using a language loss. At test time, only the scene branch is needed for sentence generation.}
\label{fig:overall_arch}
\vspace{-10pt}
\end{figure*}

\section{Related Work}

\noindent\textbf{General Video Classification.}
Spatio-temporal reasoning is one of the main topics for video understanding. With the success of deep Convolutional Neural Networks (CNNs) on image recognition \cite{NIPS2012AlexNet}, many deep architectures have been proposed correspondingly in the space-time domain. C3D \cite{ICCV2015C3D} and I3D \cite{CVPR2017I3D} construct hierarchical spatio-temporal understanding by performing 3D convolution. The two-stream network \cite{CVPR2016TwoStream} receives additional motion information by fusing an extra optical flow branch. TSN \cite{ECCV2016TSN}, on the other hand, takes advantage of the fact that huge redundancy exists between adjacent video frames via sparse frame sampling. While arguing that previous methods fail to capture long-term dependency, several recent works \cite{Arxiv2018SlowFast, CVPR2018NonLocal, CVPR2019FeatureBank, CVPR2018TRN} attempt to model a wider temporal range. Specifically, TRN \cite{CVPR2018TRN} extends TSN by considering multi-level sampling frequency. The non-local network \cite{CVPR2018NonLocal} explicitly creates long-term spatio-temporal links among features. The SlowFast network \cite{Arxiv2018SlowFast} exploits multiple time scales by creating two pathways with different temporal resolutions. Alternatively, the long-term feature bank \cite{CVPR2019FeatureBank} directly stores long-term features and later correlates them with short-term features. However, all these models directly reason over raw pixels, which often fail to ground their predictions to visual evidence by simply collecting data bias. In contrast, we propose to model relationships over higher-level entities, which in our case, are the objects within scenes.

\noindent\textbf{Spatio-Temporal Graphs.}
While the idea of graphical scene representation has been explored extensively in the image domain \cite{CVPR2015ImgRetSceneGraph, IJCV2017VG, ECCV2018GraphRCNN}, its extension to videos has only been recently attracting attention. Among the earlier attempts, ST-GCN \cite{AAAI2018STGCN} models human body joint coordinates to perform action classification. Later works directly model the objects in a scene. The resulting representation is then used to perform various down-stream tasks, such as action classification \cite{Arxiv2018CollisionSTGraph, ECCV2018STRegionGraph, CVPR2019GroupActivityGraph}, action localization \cite{Arxiv2018StackedSTGraph, Arxiv2019SymbolicSTGraph}, relation prediction \cite{CVPR2019GatedSTGraph}, and gaze prediction \cite{Arxiv2019GazeSTGraph}. All these works aim for simple classification or localization tasks where capturing object interactions might not be as important. Thus the effect of spatio-temporal graph remains unclear. In this work, we target at the much harder task of video captioning, and show the efficacy of our graph-based approach through extensive experiments and ablation study. While previous methods suffer from the noisy feature learning problem, we solve it via the proposed object-aware knowledge distillation mechanism.

\noindent\textbf{Knowledge Distillation.}
Knowledge distillation was first proposed in \cite{Arxiv2015KnowledgeDistill}, where the distillation is performed from a large model to a small one by minimizing the KL divergence between their logits distributions. Later, Lopez-Paz \etal~\cite{Arxiv2015UnifyDistill} generalize distillation to incorporate privileged information, which is some additional information that is available during training but not accessible during testing. One application of this approach is to treat the extra modality as the privileged information \cite{CVPR2016CrossModalDistill}. In our case, we innovatively regard object interactions as the privileged information. We leverage such information during training by distilling it into the scene branch, while only the scene branch is executed during testing.

\noindent\textbf{Video Captioning.}
Earlier work on video captioning mainly focus on template-based language models \cite{ICCV2013Youtube2Text, GCPR2014CoherentDescription, ICCV2013Translating}. Motivated by the success of the encoder-decoder architecture, Venugopalan \etal~\cite{Arxiv2014EncDec} extend it to the field of video captioning by globally pooling all frame features. The following works then try to exploit temporal patterns by introducing attention mechanisms \cite{ECCV2018PickNet, ICCV2015S2VT}. Very recently, Pei \etal~\cite{CVPR2019MARN} propose MARN, which attends to all semantically similar videos when generating descriptions for a single video. Wang \etal~\cite{Arxiv2019ControllableVideoCap} and Hou \etal~\cite{ICCV2019JointSyntaxVideoCap} provide the idea of predicting POS information before the actual sentence. While Recurrent Neural Networks (RNNs) are adopted as the language decoder for most of the models, Transformer \cite{NIPS2017Transformer} has been shown to be powerful as well \cite{ACML2018TVT, CVPR2019GroundedVideoCap, CVPR2018DenseCap}. Because it is faster and easier to train, we employ Transformer as the language decoder in our model.

Although most of the prior work directly operates on the global frames or video features, there have been a few attempts that try to model local object features. Zhou \etal~\cite{CVPR2019GroundedVideoCap} and Ma \etal~\cite{CVPR2018AttendNInteract} both use spatial pooling to aggregate object features. Zhang \etal~\cite{CVPR2019OABTG} propose to perform object tracking and model object trajectories using GRU. However, they either ignore the temporal \cite{CVPR2018AttendNInteract, CVPR2019GroundedVideoCap} or the spatial \cite{CVPR2019OABTG} object interactions. We instead model both spatial and temporal object interactions jointly via our proposed spatio-temporal graph. Moreover, our approach is able to incorporate prior knowledge into the adjacency matrix, which provides better interpretability than the fully learned attention mechanism.

\section{Method}

An overview of our proposed two-branch network architecture is illustrated in Fig.~\ref{fig:overall_arch}. During the training process, given a video that depicts a dynamic scene, our goal is to condense it into a representation that fully captures the spatio-temporal object interaction. This is done via the proposed spatio-temporal graph network, which serves as the object branch. Afterward, this interaction information is distilled into another scene branch via the object-aware knowledge distillation mechanism. At test time, only the scene branch is retained to generate text descriptions. In the following, we will describe each part in detail.

\subsection{Feature Representation}
Given a sequence of RGB frames $\{x_1, x_2, \ldots, x_T\}$, we extract two types of features out of them: scene features and object features. 

\noindent\textbf{Scene Features.} We follow the procedure in \cite{CVPR2019MARN}, where we first extract a sequence of 2D frame features $F_{2D} = \{f_1, f_2, \ldots, f_T\}$ using ResNet-101 \cite{CVPR2016ResNet}, with each $f_t \in \mathbb{R}^{d_{2D}}$. We also extract a set of 3D clip features $F_{3D} = \{v_1, v_2, \ldots, v_L\}$ using I3D \cite{CVPR2017I3D}, where $v_l \in \mathbb{R}^{d_{3D}}$.

\noindent\textbf{Object Features.} We run Faster R-CNN \cite{ICCV2015FasterRCNN} on each frame to get a set of object features $F_{o} = \{o_1^{1}, o_1^{2}, \ldots, o_t^{j}, \ldots, o_{T}^{N_{T}}\}$, where $N_{t}$ denotes the number of objects in frame $t$ and $j$ is the object index within each frame. Each $o_t^{j}$ has the same dimension $d_{2D}$ as $F_{2D}$.

\subsection{Spatio-Temporal Graph}

Objects have radically different behaviors across the space and time domains. On the one hand, different objects interact with each other spatially. While on the other hand, the same objects transform (shape, location, pose, etc.) temporally. In order to capture these two types of correlations, we decompose our graph into two components: the spatial graph and the temporal graph. A unique undirected spatial graph is instantiated for each frame, whose adjacency matrix is denoted by $G_t^{space}$ for time step $t$. For the temporal graph, in order to not overwhelm the model with noisy information, we only calculate temporal edges between an adjacent frame pair instead of in a fully-connected manner \cite{Arxiv2018StackedSTGraph, ECCV2018STRegionGraph}. Note that the temporal graph is still connected across all time steps in this way. The resulted temporal graph going from $t$ to $t+1$ is represented as $G_{t}^{time}$, which is a directed graph following along the direction of time flow.

\noindent\textbf{Spatial Graph.}
The goal of the spatial graph is to capture interactions among spatially related objects. Take the scene shown in Fig.~\ref{fig:overall_arch} for example. With the help of the object detector, we know there is a ``cat" as well a ``box" in the scene, but how can we get a clue on whether the cat is interacting with the box? The crux of solving this problem lies in the relative spatial location of the objects. Based on the observation that objects which are close to each other are more likely to be correlated, we explicitly incorporate this information in the spatial graph by connecting objects using their normalized Intersection over Union (IoU) value:
\begin{equation}
    G_{tij}^{space} = \frac{\exp \sigma_{tij}}{\sum_{j=1}^{N_{t}} \exp \sigma_{tij}},
\end{equation}
where $G_{tij}^{space}$ is the $(i, j)$-th element of $G_t^{space} \in \mathbb{R}^{N_t \times N_t}$, which measures the spatial connectivity between the $i$th and $j$th objects at time step $t$. We adopt the Softmax function as the normalization function similar to \cite{ECCV2018STRegionGraph, CVPR2019GroupActivityGraph}. $\sigma_{tij}$ denotes the IoU between the two objects. 

\noindent\textbf{Temporal Graph.}
While the spatial graph has the capability of capturing interactions among objects at one time step, it is unable to model the object transformations over time. In the example in Fig.~\ref{fig:overall_arch}, there is no way to tell what the cat is doing with the box with any single frame. To this end, we propose to connect all semantically similar objects in every adjacent frame pair by computing their pair-wise cosine feature similarity:
\begin{equation}
    G_{tij}^{time} = \frac{\exp \cos{ ( o_{t}^{i}, o_{t+1}^{j} ) }}{\sum_{j=1}^{N_{t+1}} \exp \cos{ ( o_{t}^{i}, o_{t+1}^{j} ) }},
\end{equation}
where $G_{tij}^{time}$ denotes the $(i, j)$-th element of $G_t^{time} \in \mathbb{R}^{N_t \times N_{t+1}}$, and $\cos{(o^i, o^j)}$ measures the cosine similarity between the two feature vectors.

\noindent\textbf{Convolutions on the Spatio-Temporal Graph.}
After we get the topological graph structure following the procedure above, the next step is to update the node features based on this graph structure. We adopt Graph Convolution (GCN) \cite{Arxiv2016GCN} for this. In order to extend the original GCN to our space-time domain, we first merge all spatial and temporal graphs for a video into a single spatio-temporal graph $G^{st}$:
\begin{equation}
    \label{eq:st_mat}
    G^{st} = 
    \begin{bmatrix}
        G_{1}^{space} &     G_{1}^{time}    &   0               &   \dots   &     0             \\
        0             &     G_{2}^{space}   &   G_{2}^{time}    &   \dots   &     0             \\
        0             &     0               &   G_{3}^{space}   &   \dots   &     0             \\
        \vdots        &     \vdots          &   \vdots          &   \ddots  &     \vdots        \\     
        0             &     0               &   0               &   \dots   &     G_{T}^{space} \\
    \end{bmatrix}
    \in \mathbb{R}^{N \times N},
\end{equation}
where each $G_t^{space}$ and $G_t^{time}$ are the spatial and temporal adjacency matrices we defined above. Note that the 0s in Eq.~\ref{eq:st_mat} are zero-valued matrices, whose shapes are determined correspondingly by the neighboring space and time matrices. $N$ is the total number of objects in the video, i.e., $N = \sum_{t=1}^T N_t$.

At this point, the graph can be updated via the standard graph convolution, which is formally defined as follows:
\begin{equation}
    H^{(l+1)} = \text{ReLU} (H^{(l)} + \Lambda^{-\frac{1}{2}} G^{st} \Lambda^{-\frac{1}{2}} H^{(l)} W^{(l)} ),
\end{equation}
where $W^{(l)} \in \mathbb{R}^{d_{model} \times d_{model}}$ is the weight matrix of layer $l$. $\Lambda$ is the diagonal degree matrix with $\Lambda_{ii}=\sum_{j}G^{st}_{ij}$. We follow \cite{AAAI2018STGCN} to add in the residual connection and use ReLU as the activation function. GCN is implemented by performing $1 \times 1 \times 1$ convolution on the input tensor $H^{(l)}$ followed by multiplying the resulting tensor with $\Lambda^{-\frac{1}{2}} G^{st} \Lambda^{-\frac{1}{2}}$. $H^{(l)} \in \mathbb{R}^{N \times d_{model}}$ is the activation from layer $l$. Particularly, $H^{(0)}$ are the stacked object features:
\begin{equation}
    H^{(0)} = \text{stack}(F_{o})W_{o} \in \mathbb{R}^{N \times d_{model}},
\end{equation}
where $\text{stack()}$ stacks all object features in $F_{o}$ along the first axis, and $W_{o} \in \mathbb{R}^{d_{2D} \times d_{model}}$ is the transformation matrix. 

Then we perform spatial average pooling on the updated $H^{N_l}$ ($N_l$ is the number of graph convolution layers), after which we get the final object features as $F_{o}' \in \mathbb{R}^{T \times d_{model}}$.

\subsection{Scene Branch}
Similar to previous work \cite{Arxiv2019GazeSTGraph, Arxiv2018StackedSTGraph, CVPR2019GatedSTGraph, ECCV2018STRegionGraph, CVPR2019OABTG, CVPR2019GroundedVideoCap}, we also directly model the frame sequence through a separate scene branch. This branch provides additional global context information that may be missing from the local object features, and is especially critical when a video has no or very few objects detected. In order to highlight the effect of our proposed spatio-temporal graph and isolate the performance from the progress in scene modeling, we keep this scene branch as simple as possible. Concretely, for every 16 consecutive non-overlapping frames, we extract one 3D feature. Then we replicate the 3D features 16 times along temporal dimension (as each 3D feature spans and provides the context across 16 time steps), and sample the $T$ slices corresponding to the 2D features. Subsequently, we project 2D and 3D features to the same dimension $d_{model}$, then concatenate them together and project again to $d_{model}$:
\begin{equation}
    F_{s} = [F_{2D}W_{2D}; F'_{3D}W_{3D}]W_{fuse} \in \mathbb{R}^{T \times d_{model}},
\end{equation}
where $W_{2D} \in \mathbb{R}^{d_{2D} \times d_{model}}$, $W_{3D} \in \mathbb{R}^{d_{3D} \times d_{model}}$ and $W_{fuse} \in \mathbb{R}^{2d_{model} \times d_{model}}$ are transformation matrices. $F'_{3D}$ represents the 3D features after the process stated above. $[;]$ denotes concatenation along channel dimension.

\subsection{Language Decoder}
During training, we pass in both scene features $F_{s}$ and object features $F_{o}'$ to perform language decoding. At test time, only $F_{s}$ is used to generate the predicted sentence. Again as our work focuses on the visual encoding component, we keep the language decoder as simple as possible. We directly adopt the TVT architecture \cite{ACML2018TVT}. Specifically, the encoder takes a temporal sequence of features (either $F_s$ or $F_{o}'$) and produces an embedding. The decoder receives this embedding and the previous word encoding to generate the next word. To clarify our naming, we denote the original encoder-decoder Transformer structure as our language decoder. Please refer to \cite{ACML2018TVT} for further details on the language decoder. Note that we use two separate Transformers for our two branches, and train them simultaneously. We adopt the standard training procedure to minimize the language cross-entropy loss $L_{o\_lang}$ and $L_{s\_lang}$ for the object and scene branch, respectively.

\subsection{Object-Aware Knowledge Distillation}
The problem with merging two branches through feature concatenation \cite{ECCV2018STRegionGraph} or pooling \cite{CVPR2019OABTG}, or adding scene features as a separate graph node \cite{Arxiv2019GazeSTGraph, Arxiv2018StackedSTGraph, CVPR2019GatedSTGraph} is that videos (and even frames in the same video) contain a variable number of objects, and this makes the learned features very noisy. This is because by either merging or adding an extra node, it imposes hard constraints on features that come from two intrinsically different spaces. By contrast, we only apply soft regularization on language logits, which are essentially probability distributions, thus being able to ensure a robust feature learning process and leverage the object information at the same time. The way of aligning language logits can be thought of as doing late fusion of the two branches, rather than early fusion as direct feature merging does. Concretely, we follow \cite{Arxiv2015KnowledgeDistill} to minimize the KL divergence between word probability distribution from the two branches. Let $P_{o}(x)$ be the probability distribution (pre-Softmax logits) across the vocabulary $V$ from object branch and $P_{s}(x)$ be the distribution from scene branch. We minimize a distillation loss:
\begin{equation}
    L_{distill} = - \sum_{x \in V} P_{s}(x) \log \left( \frac{P_{o}(x)}{P_{s}(x)} \right).
\end{equation}
Note that we do not perform distillation by minimizing the L2 distance between features \cite{CVPR2016CrossModalDistill} as it is essentially putting hard constraints on features, and we will show through experiments that it yields inferior results.

\subsection{Training}
We freeze the scene and object feature extractors and only train the rest of the model. The overall loss function consists of three parts, i.e.:
\begin{equation}
    L = L_{o\_lang} + \lambda_{sl} L_{s\_lang} + \lambda_{d} L_{distill},
\end{equation}
where $\lambda_{sl}$ and $\lambda_{d}$ are trade-off hyper-parameters. 

\section{Experiments and Results}

We evaluate our proposed model on two challenging benchmark datasets: Microsoft Research-Video to Text (MSR-VTT) \cite{CVPR2016MSRVTT} and Microsoft Video Description Corpus (MSVD) \cite{ACL2011MSVD}. To have a comprehensive evaluation, we report numbers on four commonly used metrics: BLEU@4, METEOR, ROUGE-L, and CIDEr.

\subsection{Datasets}

\noindent\textbf{MSR-VTT.}
MSR-VTT is a widely used large-scale benchmark dataset for video captioning. It consists of 10000 video clips, each human-annotated with 20 English sentences. The videos cover a diverse set of 20 categories spanning sports, gaming, cooking, etc. We follow the standard data split scheme in previous work \cite{CVPR2019MARN, Arxiv2019ControllableVideoCap, CVPR2019OABTG}: 6513 video clips in training set, 497 in validation, and 2990 in testing.  

\noindent\textbf{MSVD.}
MSVD is another popular video description benchmark, which is composed of 1970 video clips collected from YouTube. It supports multi-lingual description by annotating each video clip with sentences from multiple languages. Following the standard practice \cite{CVPR2019MARN, Arxiv2019ControllableVideoCap, CVPR2019OABTG}, we only select those English captions, after which we get approximately 40 descriptions per video, and 1200, 100, 670 clips for training, validation and testing, respectively.

\subsection{Evaluation Metrics}
In our experiments, we evaluate the methods across all four commonly used metrics for video captioning, namely BLEU@4 \cite{ACL2002BLEU}, ROUGE-L \cite{2004ROUGE}, METEOR \cite{ACL2005METEOR}, and CIDEr \cite{CVPR2015CIDEr}. BLEU@4 measures the precision of 4-grams between the ground-truth and generated sentences. ROUGE-L computes a harmonic mean of precision and recall values on the longest common subsequence (LCS) between compared sentences. METEOR, on the other hand, uses a uni-grams-based weighted F-score and a penalty function to penalize incorrect word order, and it is claimed to have better correlation with human judgment. Finally, CIDEr adopts a voting-based approach, hence is considered to be more robust to incorrect annotations. We follow the standard practice to use the Microsoft COCO evaluation server \cite{Arxiv2015COCOEval}.

\subsection{Implementation Details}
\noindent\textbf{Feature Extractor.}
For scene features, we follow \cite{CVPR2019MARN} to extract both 2D and 3D features to encode scene information. We use the ImageNet \cite{CVPR2009ImageNet} pre-trained ResNet-101 \cite{CVPR2016ResNet} to extract 2D scene features for each frame. Specifically, we pass in a center-cropped frame patch with size $224 \times 224$, and take the output from the average pooling layer to get a flattened $F_{2D}$ with $d_{2D}=2048$. We also use the Kinetics \cite{Arxiv2017Kinetics} pre-trained I3D \cite{CVPR2017I3D} for 3D scene feature extraction, where the input is a video segment consisting of 16 consecutive frames and we take the output from the last global average pooling layer to obtain a $F_{3D}$ with $d_{3D}=1024$.

To extract object features, we first apply a Faster-RCNN (with ResNeXt-101 + FPN backbone) \cite{ICCV2015FasterRCNN} pre-trained on Visual Genome \cite{IJCV2017VG} to generate object bounding boxes for each frame. We set the confidence score threshold for a detection to be considered at $0.5$. Given the output bounding boxes, we apply RoIAlign \cite{ICCV17MaskRCNN} to extract features of the corresponding regions. Specifically, we first project the bounding boxes onto the feature map from the last convolutional layer of ResNeXt-101, then apply RoIAlign \cite{ICCV17MaskRCNN} to crop and rescale the object features within the projected bounding boxes into the same spatial dimension. This generates a $7 \times 7 \times 2048$ feature for each object, which is then max-pooled to $1 \times 1 \times 2048$.

\noindent\textbf{Hyper-parameters.}
For feature extraction, we uniformly sample 10 frames for both $F_{s}$ and $F_{o}$ (i.e., $T=10$). We set the maximum number of objects in each frame to be 5. Specifically, we take the 5 most confident detections if there are more, and do zero-padding if there are less.

For the spatio-temporal graph, we stack 3 graph convolution layers, whose input and output channel number are all $d_{model}=512$. In our language decoder, both the Transformer encoder and decoder have 2 layers, 8 attention heads, 1024 hidden dimension size, and 0.3 dropout ratio.

For the trade-off hyper-parameters in the loss function, we set $\lambda_{sl}$ and $\lambda_{d}$ to be 1 and 4, respectively. All hyper-parameters were tuned on the validation set.

\noindent\textbf{Other Details.}
We adopt Adam with a fixed learning rate of $1 \times 10^{-4}$ with no gradient clipping used. We train our models using batch size 64 for 50 epochs and apply early stopping to find the best-performed model. During testing, we use greedy decoding to generate the predicted sentences.  All our experiments are conducted on two TITAN X GPUs.  

\subsection{Experimental Results}
\noindent\textbf{Comparison with Existing Methods.}
We first compare our approach against earlier methods, including \textbf{RecNet} \cite{CVPR2018RecNet}, which adds one reconstructor on top of the traditional encoder-decoder framework to reconstruct the visual features from the generated caption, and \textbf{PickNet} \cite{ECCV2018PickNet} which dynamically attends to frames by maximizing a picking policy. We also compare to several very recent works that achieve strong performance. \textbf{MARN} \cite{CVPR2019MARN} densely attends to all similar videos in training set for a broader context. \textbf{OA-BTG} \cite{CVPR2019OABTG} constructs object trajectories by tracking the same objects through time. While these works generally focus on the encoding side, \textbf{Wang \etal}~\cite{Arxiv2019ControllableVideoCap} and \textbf{Hou \etal}~\cite{ICCV2019JointSyntaxVideoCap} focus on the language decoding part and both propose to predict the POS structure first and use that to guide the sentence generation.

Note that among all these methods, we use the same scene features as MARN \cite{CVPR2019MARN}, i.e., ResNet-101 and I3D, so our method is most comparable to MARN. We also follow the standard practice \cite{CVPR2019MARN} to not compare to methods based on reinforcement learning (RL) \cite{Arxiv2019ControllableVideoCap}.

The quantitative results on MSR-VTT and MSVD are presented in Table~\ref{table:results_msr_vtt} and Table~\ref{table:results_msvd}, respectively. On MSVD, our proposed method outperforms all compared methods on 3 out of 4 metrics by a large margin. While on MSR-VTT, the performance of our model is not as outstanding. We summarize the following reasons for this: (1) MSR-VTT contains a large portion of animations, on which object detectors generally fail, thus making it much harder for our proposed spatio-temporal graph to capture object interactions in them; (2) The two very recent methods, i.e., Wang \etal~\cite{Arxiv2019ControllableVideoCap} and Hou \etal~\cite{ICCV2019JointSyntaxVideoCap} both directly optimize the decoding part, which are generally easier to perform well on language metrics compared to methods that focus on the encoding part, such as ours; (3) The more advanced features used (IRv2+I3D optical flow for Wang \etal~\cite{Arxiv2019ControllableVideoCap} and IRv2+C3D for Hou \etal~\cite{ICCV2019JointSyntaxVideoCap}) make it unfair to directly compare with them. Nonetheless, our method demonstrates a clear boost over other baselines, including the most comparable one MARN \cite{CVPR2019MARN}, as well as our own baseline, i.e., Ours (Scene), where only the scene branch is used. This manifests the effectiveness of our proposed method. 

\begin{table}[ht]
\centering
\caption{Comparison with other methods on MSR-VTT (\%). ``-" means number not available. The first section includes methods that optimize language decoding, while the second is for those that focus on visual encoding.}
\resizebox{\linewidth}{!}{%
\setlength{\tabcolsep}{3pt}
\begin{tabular}{l|c|c|c|c}
\Xhline{1pt}
{Method} & {BLEU@4} & {METEOR} & {ROUGE-L} & {CIDEr} \\ 
\hline
% \hline
Wang \etal~\cite{Arxiv2019ControllableVideoCap} & 42.0 & 28.2 & 61.6 & 48.7 \\
Hou \etal~\cite{ICCV2019JointSyntaxVideoCap} & \textbf{42.3} & \textbf{29.7} & \textbf{62.8} & \textbf{49.1} \\
\hline
RecNet \cite{CVPR2018RecNet} & 39.1 & 26.6 & 59.3 & 42.7 \\
PickNet \cite{ECCV2018PickNet} & 41.3 & 27.7 & 59.8 & 44.1 \\
OA-BTG \cite{CVPR2019OABTG} & \textbf{41.4} & 28.2 & - & 46.9 \\
MARN \cite{CVPR2019MARN} & 40.4 & 28.1 & 60.7 & \textbf{47.1} \\
Ours (Scene only) & 37.2 & 27.3 & 59.1 & 44.6 \\
Ours & 40.5 & \textbf{28.3} & \textbf{60.9} & \textbf{47.1} \\
\Xhline{1pt}
\end{tabular}
}
\label{table:results_msr_vtt}
\end{table}

\begin{table}[ht]
\vspace*{2mm}
\centering
\caption{Comparison with other methods on MSVD (\%).}
\resizebox{\linewidth}{!}{%
\setlength{\tabcolsep}{3pt}
\begin{tabular}{l|c|c|c|c}
\Xhline{1pt}
{Method} & {BLEU@4} & {METEOR} & {ROUGE-L} & {CIDEr} \\ 
\hline
Wang \etal~\cite{Arxiv2019ControllableVideoCap} & 52.5 & 34.1 & 71.3 & 88.7 \\
Hou \etal~\cite{ICCV2019JointSyntaxVideoCap} & 52.8 & 36.1 & 71.8 & 87.8 \\
\hline
RecNet \cite{CVPR2018RecNet} & 52.3 & 34.1 & 69.8 & 80.3 \\
PickNet \cite{ECCV2018PickNet} & 52.3 & 33.3 & 69.6 & 76.5 \\
OA-BTG \cite{CVPR2019OABTG} & \textbf{56.9} & 36.2 & - & 90.6 \\
MARN \cite{CVPR2019MARN} & 48.6 & 35.1 & 71.9 & 92.2 \\
Ours & 52.2 & \textbf{36.9} & \textbf{73.9} & \textbf{93.0} \\
\Xhline{1pt}
\end{tabular}
}
%\vspace{-5pt}
\label{table:results_msvd}
\end{table}

\noindent\textbf{Ablation Study.}
At a high level, our proposed method consists of two main components: the spatio-temporal graph and the object-aware knowledge distillation. The spatio-temporal graph further contains two sub-components at a lower level, which are the spatial graph and the temporal graph. We evaluate the performance of several variants to validate the efficacy of each component. We first evaluate (1) \textbf{Scene Branch Only} where only the scene branch is used, (2) \textbf{Two Branch + Concat} where both branches are used, but the fusion of two branches is done by direct concatenation of features before passing into Transformers,  and (3) \textbf{Two Branch + L2} which minimizes the L2 distance between features for distillation. These are intended to show the effectiveness of the two high-level components. In order to test different types of graph connection, we evaluate the performance of (4) \textbf{Spatial Graph Only} which only calculates the spatial graph $G^{space}$ while setting $G^{time}$ to all 0s, (5) \textbf{Temporal Graph Only} which similarly constructs only the temporal graph $G^{time}$ and puts $G^{space}$ to all 0s, as well as (6) \textbf{Dense Graph} which densely connects all objects with uniform weights (i.e., $G^{st}$ is set to all 1s). (6) is also the method proposed in Wang \etal~\cite{ECCV2018STRegionGraph}. Note that we also compare with the spatial attention approach introduced in Ma \etal~ \cite{CVPR2018AttendNInteract} and Zhou \etal~\cite{CVPR2019GroundedVideoCap}, which is essentially equivalent to \textbf{Spatial Graph Only} because the attentive object aggregation only happens spatially and temporal modeling is done by passing the spatially attended object feature sequence into language decoder. The ablation study results on MSVD are shown in Table~\ref{table:results_ablation}.

\begin{table}[ht]
\centering
\caption{Ablation study on MSVD (\%).}
\resizebox{\linewidth}{!}{%
\setlength{\tabcolsep}{3pt}
\begin{tabular}{l|c|c|c|c}
\Xhline{1pt}
{Method} & {BLEU@4} & {METEOR} & {ROUGE-L} & {CIDEr} \\ 
\hline
% Object Only
Scene Branch Only & 45.8 & 34.3 & 71.0 & 86.0 \\
Two Branch + Concat & 45.5 & 34.1 & 70.7 & 79.3 \\
Two Branch + L2 & 46.1 & 33.7 & 70.6 & 80.3 \\
\hline
Spatial Graph Only & 50.8 & 36.1 & 72.9 & 91.8 \\
Temporal Graph Only & 50.7 & 36.1 & 73.1 & 92.1 \\
Dense Graph & 51.4 & 35.9 & 72.8 & 91.3 \\
\hline
Our Full Model & \textbf{52.2} & \textbf{36.9} & \textbf{73.9} & \textbf{93.0} \\
\Xhline{1pt}
\end{tabular}
}
\label{table:results_ablation}
\end{table}

We first investigate the effect of the two high-level components. Both ``Two Branch + Concat" and ``Two Branch + L2" perform worse than the ``Scene Branch Only" baseline, which suggests that imposing hard constraints on features not only fails to exploit useful object-level information, but even hurts performance by overwhelming the model with noisy features. Once making the object branch regularize the learning of the scene branch via logit alignment (which is ``Our Full Model"), the object-level information becomes useful and gives a significant performance boost. Then we analyze the role each sub-graph plays. ``Spatial Graph Only" and ``Temporal Graph Only" achieve similar results, but are both inferior to ``Our Full Model." This validates that both sub-graphs capture important and distinct information. Finally, we would like to see how much effect prior knowledge has when creating the graph. We see a large performance margin between ``Dense Graph" and ``Our Full Model," which corroborates our argument that prior knowledge about spatial layout and temporal transformation provides the model with more helpful information.

\begin{figure*}[t]
\begin{center}
\begin{tabular}{c c}
    \includegraphics[width=0.502\linewidth]{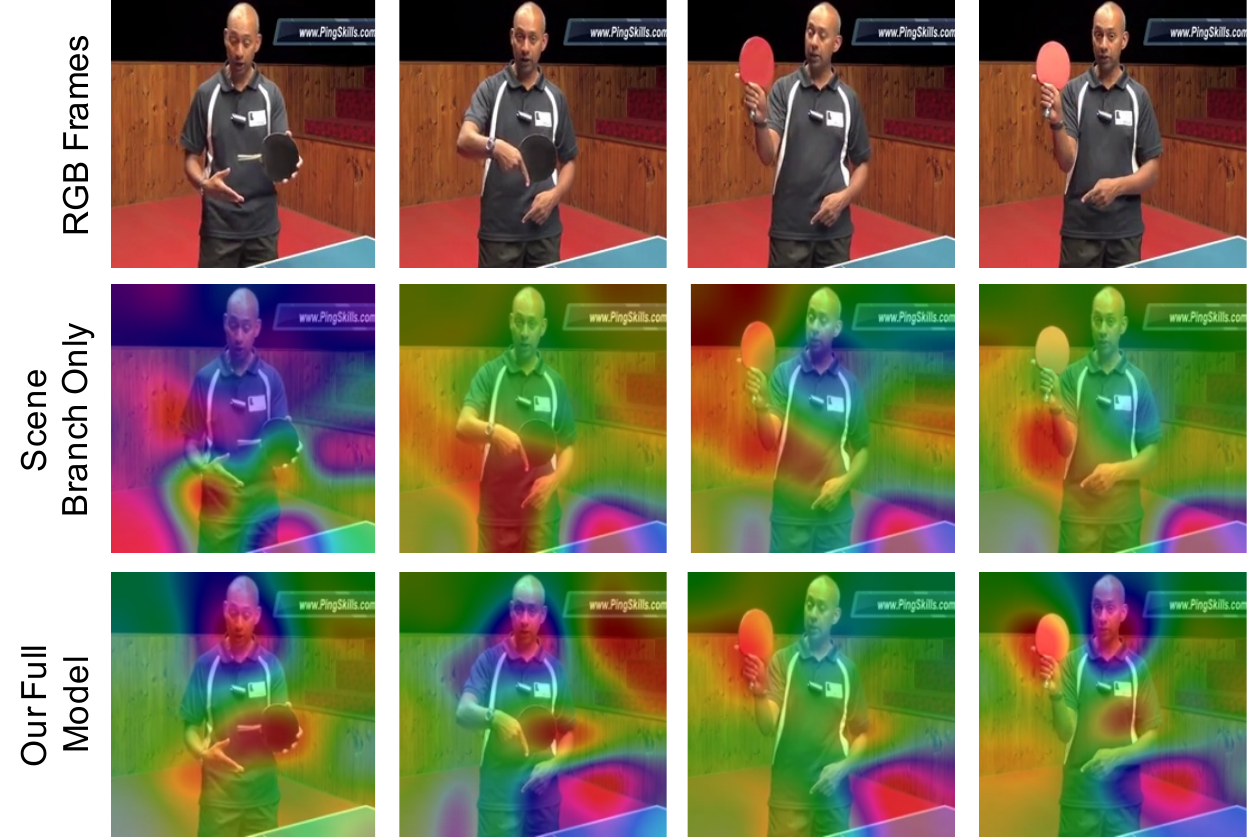} &      \includegraphics[width=0.458\linewidth]{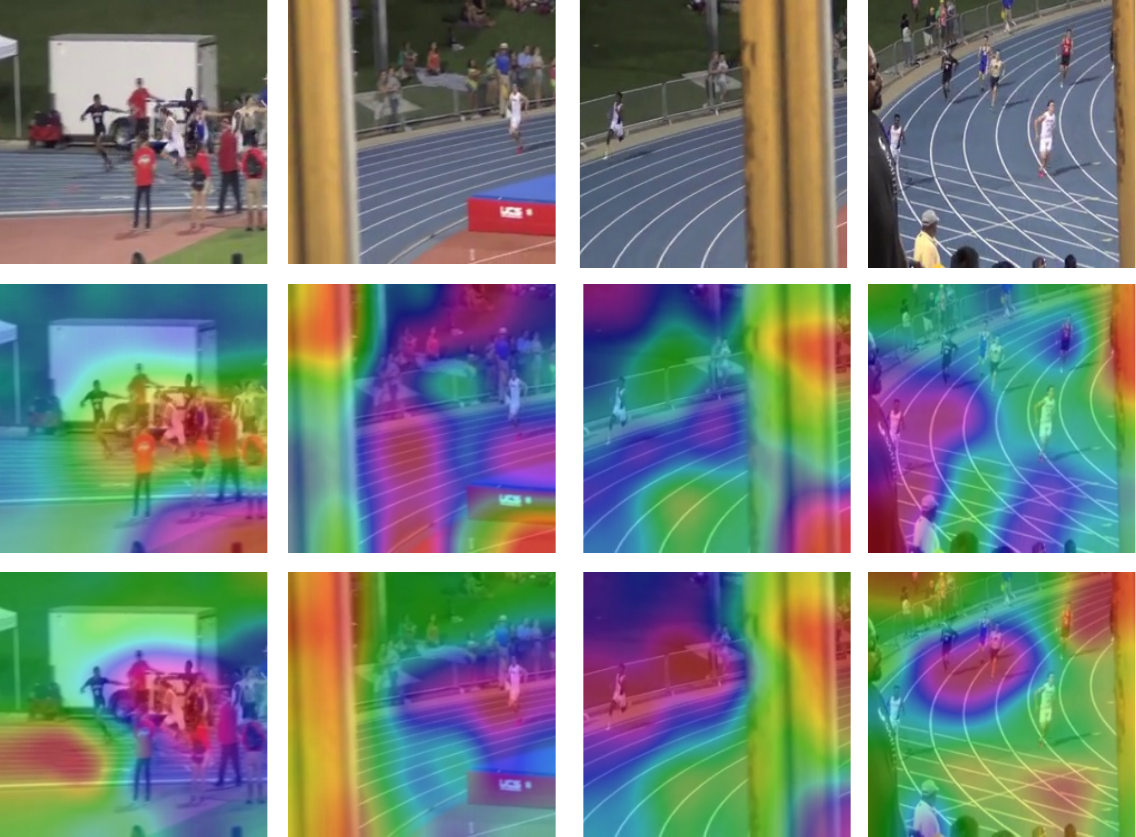} \\
    \small{GT: \footnotesize\textsf{a man in a \underline{black shirt} demonstrates how to play ping pong}} & \small{GT: \footnotesize\textsf{a group of men are running down a \underline{race} track}} \\
    \small{Wang \etal~\cite{Arxiv2019ControllableVideoCap}: \footnotesize\textsf{there is a man is talking about table tennis}} & \small{Wang \etal~\cite{Arxiv2019ControllableVideoCap}: \footnotesize\textsf{there is a man running on the track}} \\
    \small{Ours: \footnotesize\textsf{a man in a \textbf{black shirt} is talking about ping pong}} & \small{Ours: \footnotesize\textsf{a \textbf{race} is going on the track}} \\
    
    \includegraphics[width=0.502\linewidth]{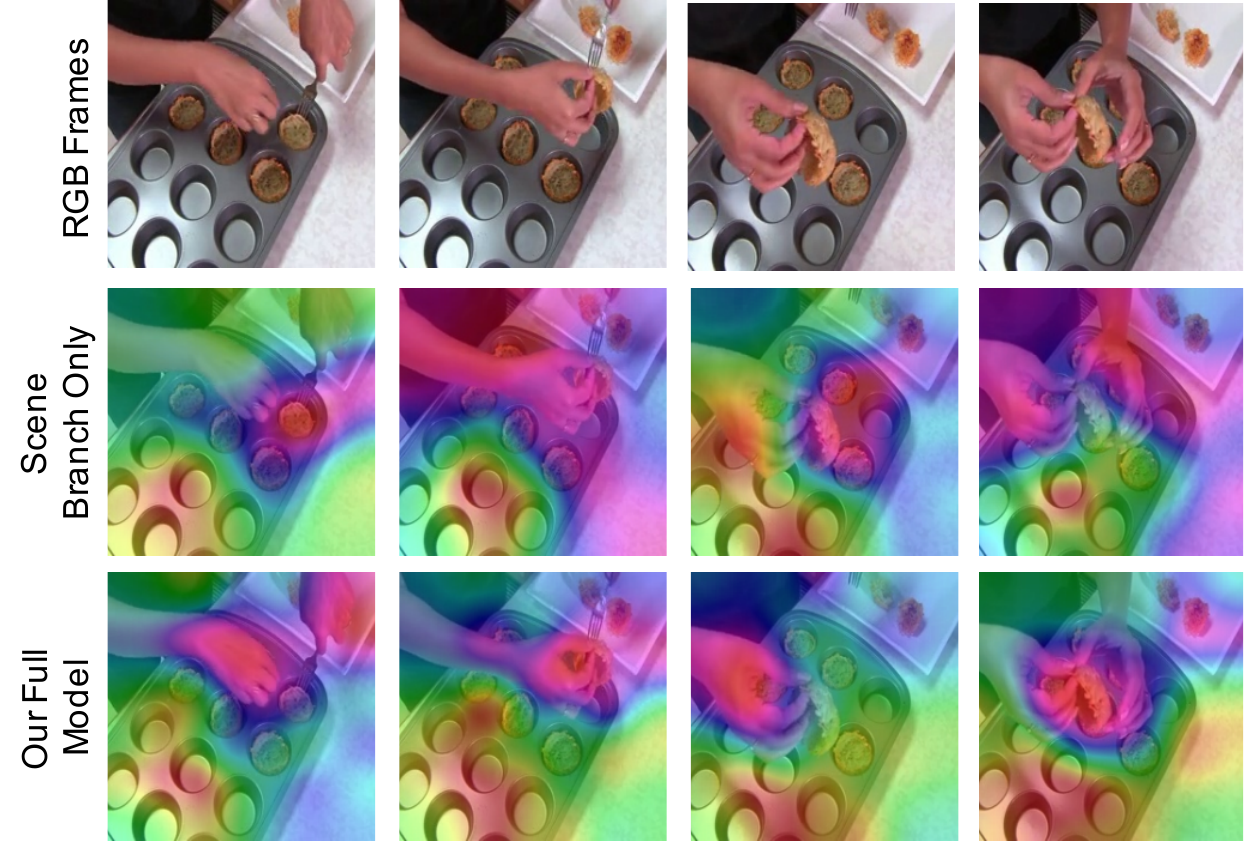} &      \includegraphics[width=0.458\linewidth]{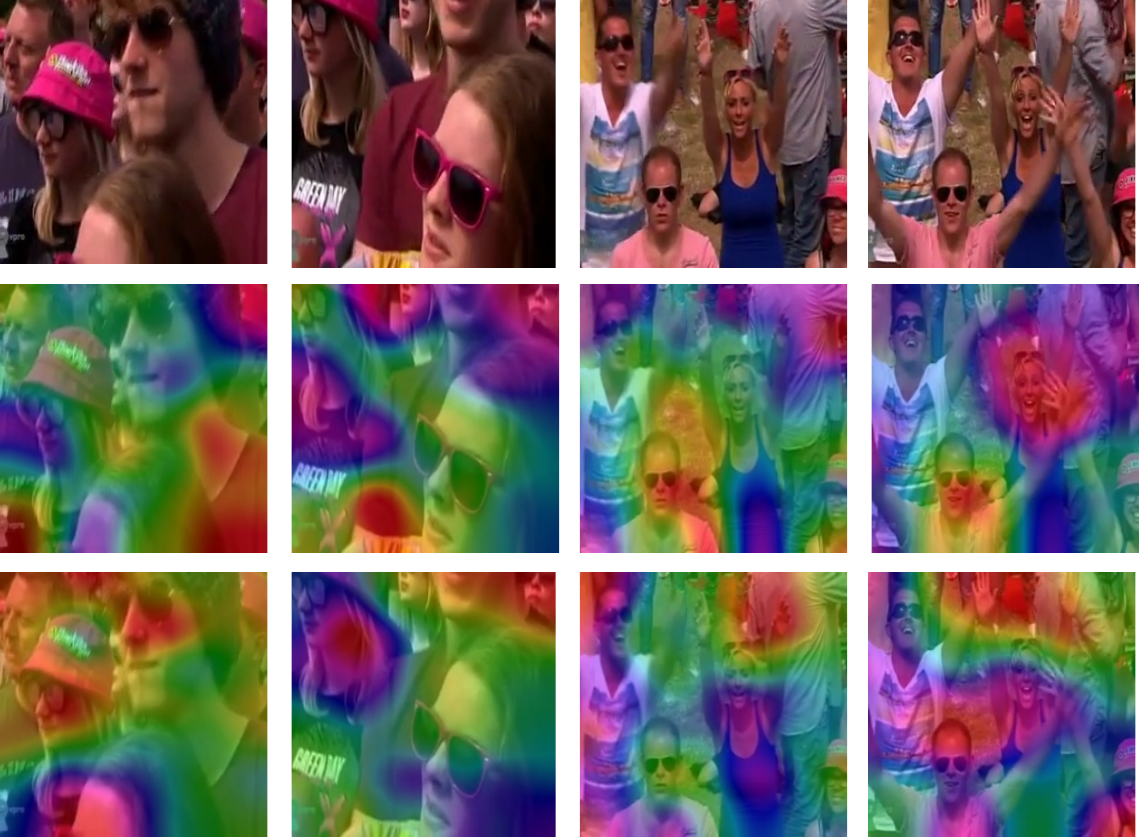} \\
        \small{GT: \footnotesize\textsf{a woman is showing how to make little baskets from \underline{potatoes}}} & \small{GT: \footnotesize\textsf{\underline{people} are dancing and singing}} \\
    \small{Wang \etal~\cite{Arxiv2019ControllableVideoCap}: \footnotesize\textsf{a person is preparing a recipe}} & \small{Wang \etal~\cite{Arxiv2019ControllableVideoCap}: \footnotesize\textsf{a man is singing}} \\
    \small{Ours: \footnotesize\textsf{a woman is showing how to make a \textbf{potato} salad}} & \small{Ours: \footnotesize\textsf{\textbf{a group of people} are singing and dancing}} \\
\end{tabular}
\vspace{-5mm}
\end{center}
   \caption{Qualitative results on 4 videos from MSR-VTT. (1) For each video, the first row shows its RGB frames, while the second and third rows are the saliency maps from our ``Scene Branch Only" and ``Our Full Model" variants (refer to ``Ablation Study" for details), respectively. Specifically, red color indicates high attention scores, while blue means the opposite. We also present the ground-truth (GT), predicted sentences from both Wang \etal~\cite{Arxiv2019ControllableVideoCap} and ``Our Full Model" (Ours).}
\label{fig:qual}
\vspace{-6pt}
\end{figure*}

\noindent\textbf{Qualitative Analysis.}
In order to validate that after distilling knowledge from the object branch our model can indeed perform better visual grounding, we plot the saliency maps for 4 example videos from MSR-VTT. Concretely, we plot for both ``Scene Branch Only" and ``Our Full Model" for comparison. We also compare the captions generated by ``Our Full Model" and Wang \etal.~\cite{Arxiv2019ControllableVideoCap}. We merge them together into Fig.~\ref{fig:qual}.

We first observe that ``Our Full Model" is able to attend to key regions much better than its ``Scene Branch Only" counterpart. In the video at the top left corner, ``Our Full Model" pays most of its attention to the man's face as well as the paddles, while ``Scene Branch Only" rarely focuses on these key parts. Similarly, in the example at the top right corner, ``Our Full Model" always keeps its attention to the group of people that are running, while the attention of ``Scene Branch Only" is mostly diffused. This further proves that our proposed spatio-temporal graph, along with the object-aware knowledge distillation mechanism, grants the model better visual grounding capability.

We then compare the captions generated from ``Our Full Model" with those from Wang \etal~\cite{Arxiv2019ControllableVideoCap}. The captions from ``Our Full Model" are generally better visually grounded than Wang \etal~\cite{Arxiv2019ControllableVideoCap}. For example, our model is able to predict very fine-grained details such as ``black shirt" for the video at the top left corner, and ``potato" for the video at the bottom left corner. It is also capable of grounding larger-scale semantic concepts, e.g., ``race" (which indicates there is more than one person) for the top-right-corner video and ``a group of people" for the bottom-right-corner one. 

\section{Conclusion}
In this paper, we propose a novel spatio-temporal graph network for video captioning to explicitly exploit the spatio-temporal object interaction, which is crucial for scene understanding and description. Additionally, we design a two-branch framework with a proposed object-aware knowledge distillation mechanism, which solves the problem of noisy feature learning present in previous spatio-temporal graph models. We demonstrate the effectiveness of our approach on two benchmark video captioning dataset.

\paragraph{Acknowledgements}
Toyota Research Institute (TRI) provided funds to assist the authors with their research, but this article solely reflects the opinions and conclusions of its authors and not TRI or any other Toyota entity. We thank our anonymous reviewers, Andrey Kurenkov, Chien-Yi Chang, and Ranjay Krishna, for helpful comments and discussion.

\newpage

{\small
\bibliographystyle{ieee_fullname}
\bibliography{egbib}
}

\end{document}